\documentclass[10pt,a4paper]{article}

\usepackage[T1]{fontenc}
\usepackage{newtxtext,newtxmath}
\usepackage[left=44mm,right=44mm,top=44mm,bottom=50mm]{geometry}
\usepackage{microtype}
\usepackage{graphicx}
\usepackage{booktabs}
\usepackage{array}
\usepackage{multirow}
\usepackage{enumitem}
\usepackage{caption}
\usepackage{url}

\title{\bfseries NLPCC 2026 Task 10: Citation-Level Faithfulness\\
Verification with DeBERTa Ensembles and Class-Wise\\
Calibration}
\author{\small Yanling Li\textsuperscript{\dag,*}\orcidlink{0009-0003-0866-0898},
Zirui Li\textsuperscript{\dag}\orcidlink{0009-0003-7904-2405}, and
Mingyu Wan\orcidlink{0000-0003-0083-5895}\\[0.8em]
\small The Hong Kong Polytechnic University, Hong Kong SAR, China\\
\fontsize{5.5pt}{6pt}\selectfont \texttt{lynnn.li@connect.polyu.hk, sebastian.li@connect.polyu.hk, mingyu.wan@polyu.edu.hk}\\
\small \textsuperscript{\dag} Equal contribution; \textsuperscript{*} Corresponding author}
\date{}

\newcommand{\orcidlink}[1]{\textsuperscript{\mbox{\fontsize{3.8pt}{4pt}\selectfont[#1]}}}

\begin{document}
\fontsize{9pt}{10.5pt}\selectfont
\emergencystretch=2em
\maketitle
\vspace{-1.5em}

\begin{abstract}
\noindent\textbf{Abstract.}
This paper presents our system for Track 2 of the NLPCC 2026 Shared Task 10 on citation-level faithfulness in AI-assisted scientific reporting. Given an atomic scientific claim and the structured full text of its cited paper, the task requires both a four-way relation label and up to three evidence paragraph identifiers. The label head ensembles a paragraph-aware cross-encoder with a document-level DeBERTa-large classifier, followed by class-wise decision calibration. Probability-level fusion is motivated by an out-of-fold tendency to over-predict Topical Match. The evidence head combines paragraph scores from top-20 and top-30 joint models with BM25 scores. The system runs fully offline without external retrieval or LLM prompting. On the final leaderboard, our system achieved 82.9898 overall (89.5491 Macro-F1 and 76.4305 Joint@3), ranking second in Track 2. Ablations and error analysis show that model complementarity and calibration drive the label gains. Gold-evidence inference changes label Macro-F1 negligibly, whereas evidence ranking remains important for Joint@3.

\medskip
\noindent\textbf{Keywords:} scientific claim verification, citation faithfulness, evidence retrieval, DeBERTa, calibration
\end{abstract}

\section{Introduction}
Generative systems increasingly assist with literature review and scientific writing, but fluent claims can exceed, distort, or merely resemble the evidence in a cited source. This failure mode is more subtle than open-domain factuality: the relevant source is already specified, yet the system must distinguish direct support from partial overstatement, topical relatedness, and irrelevance, and identify the paragraphs that justify its decision. These requirements couple document retrieval with fine-grained semantic verification.

NLPCC 2026 Shared Task 10 formalizes this setting in two tracks. Track 2 evaluates citation-faithfulness judgments for scientific claims and requires both a relation label and evidence paragraphs under an offline constraint~\cite{nlpcc2026task10}.

\clearpage

Our system uses parallel label and evidence outputs with a shared paragraph encoder. A paragraph-aware DeBERTa cross-encoder pools claim-paragraph representations into a document decision, while a DeBERTa-large classifier supplies complementary document-level probabilities. Class-wise decision calibration corrects a systematic out-of-fold bias, and evidence scores are fused across candidate windows with BM25.

\textbf{Contributions.} (1) A two-head offline architecture combining paragraph evidence and document-level supervision; (2) probability ensembling with class-wise decision calibration; (3) systematic ablations of input design, fusion, calibration, and evidence weighting; and (4) an 82.99 overall Track 2 submission.

\section{Related Work}
Automated claim verification commonly decomposes into evidence retrieval and entailment-style classification. FEVER introduced large-scale verification against Wikipedia with supporting sentences~\cite{thorne2018fever}. Scientific claim verification adds specialized terminology, longer sources, and evidential conventions: SciFact pairs expert-written claims with supporting or refuting rationales~\cite{wadden2020scifact}, while MultiVerS jointly predicts document labels and sentence rationales from full-document context~\cite{wadden2022multivers}. Our joint model follows this multi-task design, whereas the shared task uses four graded citation relations instead of a support/refute/no-evidence taxonomy.

Full-paper reasoning has been studied in QASPER, where information-seeking questions require evidence from research papers~\cite{dasigi2021qasper}, and is enabled by structured scientific resources such as S2ORC~\cite{lo2020s2orc}. Domain-specific encoders such as SciBERT improve scientific NLP through in-domain pretraining~\cite{beltagy2019scibert}. We use DeBERTa-v3~\cite{he2023debertav3} with classical BM25 retrieval~\cite{robertson1995okapi}, pairing a paragraph-aware cross-encoder with a complementary document classifier. Post-hoc decision adjustment is well motivated for modern neural classifiers~\cite{guo2017calibration}; our class-wise coefficients target the Macro-F1 decision boundary instead of calibrated posterior probabilities.

\section{Task, Data, and Evaluation}
\subsection{Task Definition}
For each instance, the input is a claim $c$ and a cited paper $D=\{p_1,\ldots,p_n\}$. The output consists of a label $y$ and up to three paragraph identifiers. Table~\ref{tab:labels} summarizes the relation labels. For non-Irrelevant instances, the evidence list contains the three highest-ranked paragraph identifiers; for an Irrelevant prediction, the official match-empty rule requires an empty evidence list.

\clearpage

\begin{table}[htbp]
\caption{Track 2 citation-faithfulness labels.}
\label{tab:labels}
\centering
\small
\begin{tabular}{@{}p{0.24\textwidth}p{0.66\textwidth}@{}}
\toprule
Label & Operational meaning \\
\midrule
Supported & The cited paper directly supports the claim. \\
Overstate & The claim extends or strengthens what the paper establishes. \\
Topical Match & The paper is topically related but does not provide the claimed evidence. \\
Irrelevant & The paper does not substantively concern the claim. \\
\bottomrule
\end{tabular}
\end{table}

\subsection{Data and Metrics}
We use only the official data: 2,162 labeled train-development instances, 937 Phase 1 test instances, and 489 hidden Phase 2 instances. No external training data or augmentation is used. The train-development label distribution is imbalanced: 1,029 Supported, 375 Overstate, 376 Topical Match, and 382 Irrelevant instances.

Label quality is measured by the unweighted mean of the four class-level F1 values. With class set $C$,
\begin{equation}
\operatorname{Macro\text{-}F1}=\frac{1}{4}\sum_{k\in C}F_1(k).
\end{equation}
Joint@3 credits a non-Irrelevant instance only when the label is correct and the predicted top-three evidence set intersects the gold evidence set. For an Irrelevant instance, credit requires both a correct Irrelevant label and an empty predicted evidence list (the official match-empty rule). The Track 2 score is the arithmetic mean of Macro-F1 and Joint@3. Final leaderboard values average the two evaluation phases. Our submitted files retained three evidence identifiers for every predicted class rather than clearing them after an Irrelevant prediction; the official scores in Table~\ref{tab:official} therefore already reflect any penalty caused by this output-rule mismatch.

\section{System}
\begin{figure}[htbp]
\centering
\includegraphics[width=\textwidth]{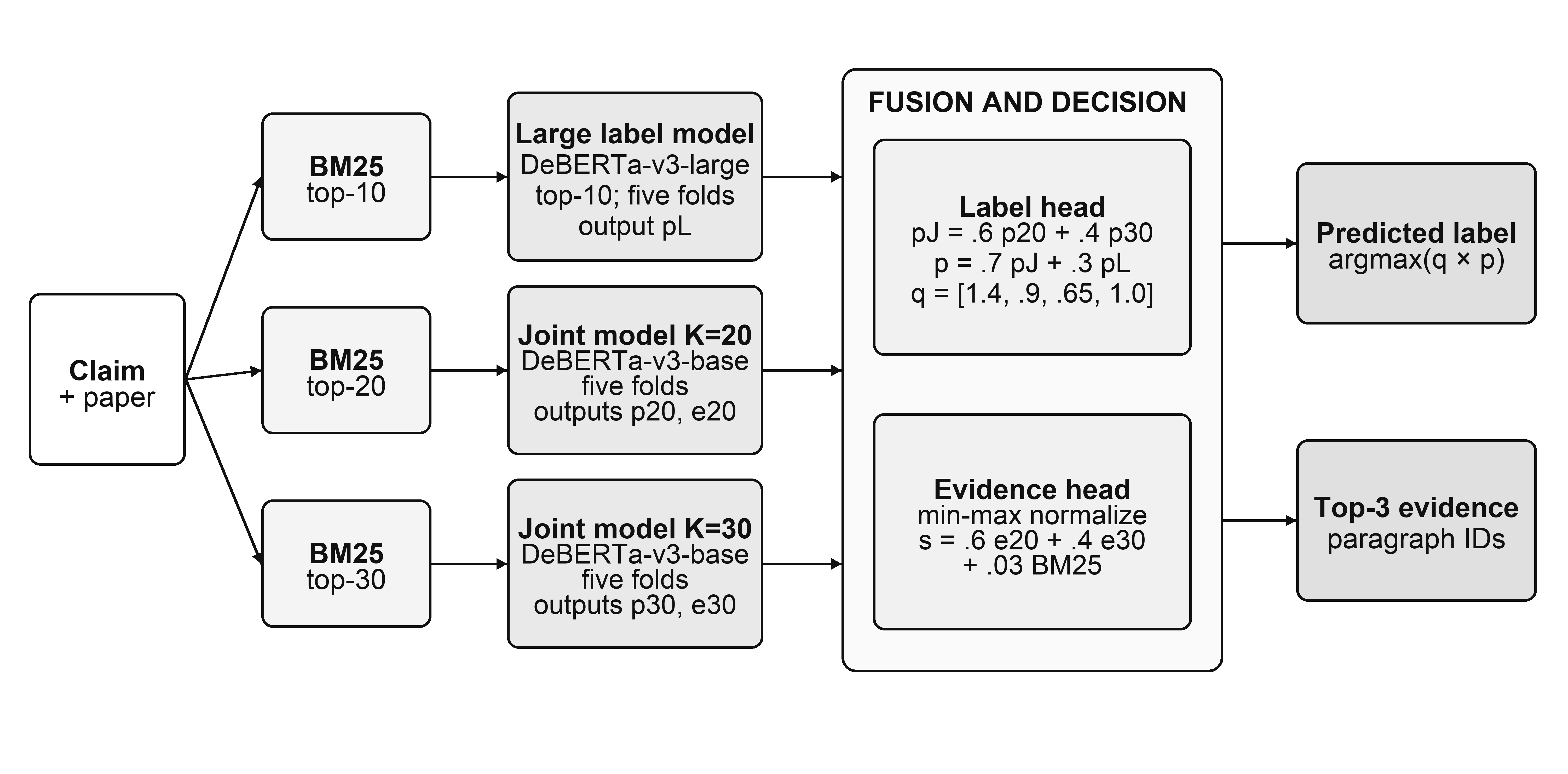}
\caption{FULL2 offline inference pipeline. The top-20 and top-30 joint systems are separately trained five-fold ensembles; the label-only large system is another five-fold ensemble.}
\label{fig:pipeline}
\end{figure}

Figure~\ref{fig:pipeline} presents the complete system. We first parse the paper into ordered paragraphs and retrieve candidates locally with BM25. Two model families then produce complementary label probabilities, while paragraph evidence scores are fused independently for ranking.

\clearpage

\subsection{Paragraph Parsing and BM25 Retrieval}
The official \texttt{cited\_paper\_full\_text} field is an ordered list of one-key paragraph dictionaries. We strip the paragraph identifier and text of surrounding whitespace and retain every non-empty pair $(p_j,\text{text}_j)$ in source order. Training-set construction uses a local BM25 implementation with lower-cased ASCII alphanumeric tokens, $k_1=1.5$, and $b=0.75$. The released inference and fusion scripts use \texttt{rank\_bm25.BM25Okapi} and tokenize ASCII alphanumerics plus individual CJK characters. Thus, the two code paths are not byte-identical, and the repository scripts define the executable preprocessing. We retain separate top-20 and top-30 candidate sets for the joint model. The label-only large model receives the claim followed by the top-10 paragraph texts, each truncated to 900 characters before model tokenization.

\subsection{Joint Label-Evidence Model}
For each candidate paragraph $p_j$, a DeBERTa-v3-base cross-encoder represents the pair $[c;p_j]$ by its first-token vector $h_j$. A scalar evidence head produces $e_j$, and a four-way relation head produces $r_j$. Evidence scores define attention weights over the $K$ candidates:
\begin{equation}
a_j=\frac{\exp(e_j)}{\sum_{t=1}^{K}\exp(e_t)}.
\end{equation}

\clearpage
The sample representation concatenates the attention-weighted paragraph vector and relation distribution. A label head maps this representation to four logits:
\begin{equation}
\bar{h}=\sum_{j=1}^{K}a_jh_j,\qquad
\bar{r}=\sum_{j=1}^{K}a_j\operatorname{softmax}(r_j),
\end{equation}
\begin{equation}
p_{\mathrm{joint}}=\operatorname{softmax}\!\left(W_{\mathrm{label}}[\bar{h};\bar{r}]+b_{\mathrm{label}}\right).
\end{equation}

Training minimizes equally weighted label cross-entropy and paragraph-level binary cross-entropy: $\mathcal{L}=\operatorname{CE}(y,p_{\mathrm{joint}})+\operatorname{BCE}(z,e)$. Here $z_j$ marks whether candidate $p_j$ is annotated as evidence, so every unannotated retrieved candidate is a negative. Irrelevant examples have no positive evidence paragraphs and therefore contribute all-zero paragraph targets. The relation head has no direct loss; gradients from document-level cross-entropy propagate through the pooled relation distribution. We train otherwise identical top-20 and top-30 models and combine their five-fold label probabilities as $p_J=0.60p_{20}+0.40p_{30}$.

\textbf{Relation-head ablation.} We trained a variant that removes the paragraph-level relation head and pools only the evidence-attention-weighted paragraph representation into the document classifier. Under the same five-fold top-20 protocol, OOF Macro-F1 decreased from 92.2156 to 88.2693, a drop of 3.9463 points. This within-run comparison shows that the pooled four-way relation features are useful within the trained architecture. It does not imply that providing better candidate paragraphs at inference must yield a comparable gain.

\subsection{Large Label Model and Probability Fusion}
A DeBERTa-v3-large sequence classifier encodes the claim together with BM25 top-10 context and outputs a four-class probability vector. Its class-weighted cross-entropy uses inverse square-root class frequencies, normalized to unit mean within each training fold. Within the label-only model family, the large model achieves higher standalone OOF Macro-F1 than the base model, while the joint model contributes complementary paragraph-aware decisions. The uncalibrated ensemble is
\begin{equation}
p_{\mathrm{fuse}}=0.70p_J+0.30p_L.
\end{equation}

\subsection{Class-Wise Decision Calibration}
We implement the final calibration as a class-wise additive logit adjustment. After choosing the fusion weight, the retained notebook generated nine calibration candidates: Mild $[1.15,\allowbreak 0.95,\allowbreak 0.85,\allowbreak 1.00]$, Topical-only $[1.00,\allowbreak 1.00,\allowbreak 0.80,\allowbreak 1.00]$, Full $[1.30,\allowbreak 0.90,\allowbreak 0.70,\allowbreak 1.00]$, and FULL2--FULL7: $[1.40,\allowbreak 0.90,\allowbreak 0.65,\allowbreak 1.00]$, $[1.50,\allowbreak 0.85,\allowbreak 0.60,\allowbreak 1.00]$, $[1.40,\allowbreak 0.85,\allowbreak 0.60,\allowbreak 0.95]$, $[1.60,\allowbreak 0.85,\allowbreak 0.55,\allowbreak 1.00]$, $[1.50,\allowbreak 0.90,\allowbreak 0.60,\allowbreak 1.05]$, and $[1.45,\allowbreak 0.80,\allowbreak 0.65,\allowbreak 0.90]$. Table~\ref{tab:calibration} reports the available public scores for the staged sequence through FULL2. A tenth conservative candidate changed FULL2's Topical Match multiplier from 0.65 to 0.70 and scored 80.79. The notebook does not establish that every generated archive was uploaded, and attempts outside it were not logged systematically; we therefore do not claim a complete leaderboard-attempt count. The selected FULL2 vector was frozen for Phase 2.
\begin{equation}
\hat{y}=\arg\max_k\left[\log p_{\mathrm{fuse},k}+\log q_k\right],
\qquad q=[1.40,0.90,0.65,1.00].
\end{equation}
This adjustment increases the effective decision margin for Supported and reduces that for Topical Match. Unlike temperature scaling, it is class-specific and directly targets Macro-F1 rather than probability calibration. Figure~\ref{fig:pipeline} shows the algebraically equivalent multiplicative implementation with class factors $q$, while Eq.~(6) presents the same decision rule as additive class-wise logit adjustment.

\subsection{Evidence Fusion}
Within each instance, we independently min-max normalize top-20 evidence scores over its candidate window ($s_{20}$), top-30 scores over its candidate window ($s_{30}$), and BM25 scores over the union of their paragraph identifiers ($b$). Paragraphs absent from a neural candidate window receive $-0.15$. The evidence ranker computes
\begin{equation}
s_{\mathrm{evidence},j}=0.60s_{20,j}+0.40s_{30,j}+0.03b_j.
\end{equation}
For non-Irrelevant predictions, the ranker returns the three highest-scoring paragraph identifiers. Under the official match-empty rule, the evidence list should instead be cleared when the final label is Irrelevant. The submitted system omitted this final clearing step, as noted in Section~3.2. Label and evidence predictions are otherwise combined only after both heads complete inference.

\section{Experimental Setup}
\subsection{Training and Reproducibility}
The saved artifacts contain two fixed, label-stratified five-fold files. The top-20 joint and label-only runs use folds of sizes 434, 432, 433, 432, and 431; the separately prepared top-30 joint run uses sizes 433, 433, 432, 432, and 432. Training uses seed 42. Both partitions are at the instance level, not grouped by cited paper or related claim family; consequently, claims associated with the same source document can occur in different folds. Each model predicts every instance only from its own held-out fold; cross-system OOF fusion aligns those predictions by sample identifier, and final test probabilities average each model's five checkpoints. Table~\ref{tab:config} records the main configuration and exact pretrained checkpoints. Model selection uses validation Track 2 score for joint checkpoints and validation Macro-F1 for label-only checkpoints. The original Colab runs installed PyTorch, Transformers, NumPy, scikit-learn, and rank-bm25 without a version lockfile; exact resolved package versions were not preserved and therefore cannot be reported retrospectively.

\clearpage

\begin{table}[htbp]
\caption{Main training and inference configuration.}
\label{tab:config}
\centering
\small
\begin{tabular}{@{}p{0.22\textwidth}p{0.70\textwidth}@{}}
\toprule
Component & Configuration \\
\midrule
Joint models & \texttt{microsoft/deberta-v3-base}; $K=20/30$; max length 384; lr $2\times10^{-5}$; 1 epoch; batch 1; accumulation 4 \\
Large model & \texttt{microsoft/deberta-v3-large}; top-10; max length 512; lr $1\times10^{-5}$; 2 epochs; batch 1; accumulation 8 \\
Optimization & AdamW; weight decay 0.01; linear schedule; 6\% warmup; gradient clipping 1.0; seed 42 \\
Cross-validation & Two fixed label-stratified instance-level files: top-20/large 434/432/433/432/431; top-30 433/433/432/432/432; seed 42; neither grouped by cited paper \\
Hardware & Single GPU (Google Colab A100 or T4, depending on availability) \\
Inference & Fully offline; local checkpoints and local BM25 only \\
\bottomrule
\end{tabular}
\end{table}

\subsection{Label Ablations}
Table~\ref{tab:label-ablation} compares input construction and model scale under clean OOF evaluation. Injecting abstract/conclusion context consumes two retrieval slots and reduces Macro-F1 by 4.04 points relative to pure BM25 input. With the same top-10 input, scaling from base to large improves Macro-F1 by 1.45 points and raises every class F1 above 0.85.

\begin{table}[htbp]
\caption{Input and model-scale ablations on five-fold OOF predictions.}
\label{tab:label-ablation}
\centering
\small
\begin{tabular}{@{}lll@{}}
\toprule
Configuration & Backbone & OOF Macro-F1 \\
\midrule
Context injected, top-8 & DeBERTa-v3-base & 83.91 \\
Pure BM25, top-10 & DeBERTa-v3-base & 87.95 \\
Pure BM25, top-10 & DeBERTa-v3-large & 89.40 \\
\bottomrule
\end{tabular}
\end{table}

Fusion weights were selected before calibration. With the large-model weight swept over $\{0.08,0.10,0.20,0.30,0.50,0.70,1.00\}$, Phase 1 scores were 78.99, 79.22, 79.37, 79.39, 79.00, 78.34, and 76.71, respectively. We therefore fixed the large-model weight at 0.30 and used this mixture for the subsequent calibration study. These are public Phase 1 evaluations; they complement, but do not replace, the OOF diagnostics used to inspect the model behavior.

\subsection{Calibration Ablation}
Table~\ref{tab:calibration} isolates the effect of increasingly targeted decision calibration on Phase 1. The final coefficient vector improves Macro-F1 from 84.61 to 86.99 and the overall score by 1.89 points. Since the evidence ranking is fixed, the Joint@3 increase is caused by more correct labels, not a changed retriever.

\clearpage

\begin{table}[htbp]
\caption{Effect of class-wise calibration on the Phase 1 Track 2 score.}
\label{tab:calibration}
\centering
\small
\begin{tabular}{@{}lr@{}}
\toprule
Variant & Phase 1 score \\
\midrule
L30, no calibration & 79.39 \\
Mild calibration & 80.03 \\
Topical-only calibration & 80.32 \\
Full calibration & 80.78 \\
FULL2: $[1.4,0.9,0.65,1.0]$ & 81.28 \\
\bottomrule
\end{tabular}
\end{table}

To assess selection stability, we repeated a split-half study five times. On each search half, one greedy coordinate pass considered $\{0.50,\allowbreak 0.60,\allowbreak 0.65,\allowbreak 0.70,\allowbreak 0.75,\allowbreak 0.80,\allowbreak 0.85,\allowbreak 0.90,\allowbreak 0.95,\allowbreak 1.00,\allowbreak 1.10,\allowbreak 1.20,\allowbreak 1.30,\allowbreak 1.40,\allowbreak 1.45,\allowbreak 1.50,\allowbreak 1.60\}$ for each of the four class multipliers in label order, then evaluated on the other half. FULL2 beat the uncalibrated prediction in four of five held-out halves, with gains of approximately 0.4--1.1 Macro-F1 points in the winning trials. These post-hoc checks use fixed OOF probabilities; they are neither repeated model training nor a significance test.

\subsection{Evidence Ablation}
BM25 contributes only a small correction after neural evidence fusion. We fixed the top-20/top-30 neural weights at 0.60/0.40, then swept the BM25 multiplier over $\{0,\allowbreak 0.03,\allowbreak 0.05,\allowbreak 0.08,\allowbreak 0.10,\allowbreak 0.12,\allowbreak 0.15,\allowbreak 0.18,\allowbreak 0.20,\allowbreak 0.25,\allowbreak 0.30\}$ on OOF predictions. Table~\ref{tab:evidence-weight} reports representative values. The response is flat: 0.03 is the grid optimum, while the displayed settings fall within 0.37 Joint@3 points. We therefore fixed the evidence head before label calibration. For this OOF diagnostic, a correct Irrelevant label is counted as a joint hit; numerically, this is equivalent to applying the required empty-list post-processing before evaluation, although the raw saved outputs were not serialized that way.

\begin{table}[htbp]
\caption{BM25 evidence weight on OOF predictions.}
\label{tab:evidence-weight}
\centering
\small
\begin{tabular}{@{}lr@{}}
\toprule
BM25 weight $\alpha$ & OOF Joint@3 \\
\midrule
0.00 & 89.13 \\
0.03 & 89.27 \\
0.05 & 89.22 \\
0.10 & 89.08 \\
0.20 & 88.90 \\
\bottomrule
\end{tabular}
\end{table}

\clearpage

We audited retrieval coverage independently of model training. On the 2,162 train-development instances, BM25 top-20 contains at least one gold paragraph for 98.15\% of non-Irrelevant instances and top-30 raises this to 99.10\%; mean gold-evidence recall increases from 91.56\% to 95.82\%. The saved five-fold top-30 OOF predictions provide a complementary evidence-level audit: overall Evidence Hit@3 is 85.62\% and Joint@3 is 81.96\%. Unlike Table~\ref{tab:evidence-weight}, this diagnostic uses standalone top-30 OOF predictions rather than the full top-20/top-30 evidence-fusion pipeline, so the values are not directly comparable. Table~\ref{tab:retrieval} reports the label-conditioned view; dashes denote that Irrelevant examples have no gold evidence. The same virtual empty-list convention is used for the OOF Joint@3 column. These are train-development OOF diagnostics, not official test-set scores.

\begin{table}[htbp]
\caption{Retrieval ceiling and label-conditioned top-30 OOF evidence performance.}
\label{tab:retrieval}
\centering
\scriptsize
\begin{tabular}{@{}lrrrrrr@{}}
\toprule
& \multicolumn{4}{c}{BM25 candidate retrieval} & \multicolumn{2}{c}{Top-30 OOF prediction} \\
\cmidrule(lr){2-5}\cmidrule(lr){6-7}
Group & H@20 & H@30 & R@20 & R@30 & E-H@3 & J3 \\
\midrule
Overall & 98.15 & 99.10 & 91.56 & 95.82 & 85.62 & 81.96 \\
Supported & 97.67 & 98.74 & 90.96 & 95.55 & 78.91 & 75.12 \\
Overstate & 98.40 & 99.47 & 89.43 & 94.64 & 92.00 & 88.00 \\
Topical Match & 99.20 & 99.73 & 95.35 & 97.74 & 97.61 & 85.64 \\
Irrelevant & -- & -- & -- & -- & -- & 90.84 \\
\bottomrule
\end{tabular}
\end{table}

\section{Official Results}
Table~\ref{tab:official} reports the official final leaderboard values. The identical FULL2 architecture and coefficients were used for both phases. Phase 2 is stronger than Phase 1 on both metrics. This hidden-set improvement is consistent with, though does not by itself prove, the robustness suggested by the OOF split-half checks.

\clearpage

\begin{table}[htbp]
\caption{Official Track 2 results for our system. Final values average the two phases.}
\label{tab:official}
\centering
\small
\begin{tabular}{@{}lrrr@{}}
\toprule
Evaluation & Macro-F1 & Joint@3 & Track 2 score \\
\midrule
Phase 1 & 86.9902 & 75.5603 & 81.2753 \\
Phase 2 & 92.1080 & 77.3006 & 84.7043 \\
Final & 89.5491 & 76.4305 & 82.9898 \\
\bottomrule
\end{tabular}
\end{table}

Final Macro-F1 exceeds Joint@3 by 13.12 points, indicating stronger relation classification than evidence localization. This difference motivates the retrieval-focused limitations discussed below and cautions against interpreting a correct relation label as sufficient evidence localization.

\section{Analysis}
\subsection{Where Calibration Helps}
The raw large-model OOF confusion matrix is shown in Table~\ref{tab:confusion}. Topical Match is the largest destination for errors from Overstate and Irrelevant, while Supported errors split between Irrelevant and Topical Match. Applying the FULL2 coefficient vector to these same large-model probabilities changes only 33 of 2,162 decisions: 21 wrong predictions become correct and 9 correct predictions become wrong. Macro-F1 rises from 89.40 to 90.00. Calibration primarily affects a small set of boundary cases without altering predictions globally.

\begin{table}[htbp]
\caption{Raw DeBERTa-large OOF confusion matrix (rows: gold; columns: prediction).}
\label{tab:confusion}
\centering
\small
\begin{tabular}{@{}lrrrr@{}}
\toprule
Gold & Sup. & Over. & Top. & Irr. \\
\midrule
Supported & 968 & 11 & 19 & 31 \\
Overstate & 6 & 352 & 12 & 5 \\
Topical Match & 12 & 21 & 326 & 17 \\
Irrelevant & 10 & 14 & 32 & 326 \\
\bottomrule
\end{tabular}
\end{table}

The most frequent decision changes are Topical Match to Overstate (10 cases), Overstate to Irrelevant (9), and Topical Match to Irrelevant (5). The pattern confirms that a single shared argmax boundary is poorly matched to the class-specific precision-recall trade-offs of Macro-F1.

\subsection{Qualitative Cases}
Table~\ref{tab:cases} presents representative OOF cases selected from stored predictions. These are diagnostic examples, not additional test-set evaluation. They show both the value and the limit of global decision calibration.

\clearpage

\begin{table}[htbp]
\caption{Representative OOF decisions and diagnoses.}
\label{tab:cases}
\centering
\scriptsize
\begin{tabular}{@{}p{0.27\textwidth}p{0.65\textwidth}@{}}
\toprule
ID and decision & Claim cue and diagnosis \\
\midrule
001517 / Topical $\rightarrow$ Supported / (corrected) & The claim accurately describes NLF's two-stage transformer. Gold P4 explicitly states that one stage aggregates along epipolar lines and the next along reference views. Calibration recovers direct support. \\
000099 / Topical $\rightarrow$ Overstate / (corrected) & The source reports selected simulation and real-robot tasks, whereas the claim generalizes to diverse real-world domains and uses the cue ``consistently''; the gold label is Overstate. \\
000211 / Supported $\rightarrow$ Supported / (still wrong) & The claim transfers BYOL's online/target-network mechanism to a modality-specific masked student-teacher system. Strong local lexical overlap makes the prediction highly confident, so a global coefficient cannot repair the compositional mismatch. \\
001105 / Supported $\rightarrow$ Supported / (still wrong) & The claim redefines software parameter-efficient adapters as hardware deployment modules. Terminology and cited adapter names create a misleading topical signal; detecting the category error requires finer semantic comparison. \\
\bottomrule
\end{tabular}
\end{table}

\subsection{Evidence-Retrieval Bottleneck}
The available top-30 OOF audit reaches 81.96 Joint@3. Two oracle checks isolate different error sources. A gold-label oracle retains the predicted evidence and raises Joint@3 to 88.16\% (1906/2162), showing the contribution of label errors. The gold-evidence-input oracle instead evaluates the original top-20 joint checkpoints with annotated evidence paragraphs supplied as the inference candidates for each non-Irrelevant held-out sample; Irrelevant samples retain their normal candidates. Without retraining, Macro-F1 changes from 91.9019 to 91.9074, only $+0.0055$ points. The oracle run's ordinary reference (91.9019) and the relation-ablation run's reference (92.2156) come from separate stored top-20 runs, so we interpret only each within-run delta. The results are not contradictory: the oracle changes which paragraphs are supplied to an already trained model, whereas the ablation removes a learned four-way relation representation and changes optimization. Together, the results indicate that BM25 top-20 already provides sufficient candidate context for label classification, while the relation features help organize that context. Accurate evidence ordering remains necessary for Joint@3: BM25 candidate hit rates are already 98.15\%/99.10\%, yet top-30 Evidence Hit@3 is 85.62\%.

\section{Limitations and Compliance}
First, the saved OOF runs use two fixed instance-level partition files and one training seed. Because neither partition is grouped by cited paper, document-level overlap can make OOF estimates optimistic; the reported within-run OOF numbers should therefore be read as controlled comparisons on the stated partition, not independent-document generalization estimates. We did not repeat model training or conduct significance tests. Second, fusion and calibration were selected sequentially from OOF analyses with limited public-leaderboard checks; the split-half diagnostic reduces but does not eliminate selection bias. Third, class-wise logit adjustment targets the observed label distribution and may require re-estimation under distribution shift. Fourth, the submitted files did not implement the official empty-evidence output for Irrelevant predictions. Finally, BM25 can miss semantically relevant paragraphs, the 512-token large-model limit forces aggressive compression, and our qualitative analysis is restricted to OOF examples because official test annotations were unavailable.

All Phase 1 and Phase 2 inference is fully offline. Evidence retrieval runs locally over the provided cited paper, neural scoring uses local checkpoints, and fusion and calibration are local numerical operations. No web search, browsing, external retrieval API, or online LLM is used at prediction time. The system uses only organizer-provided task data for training and evaluation.

\section{Conclusion}
We presented a fully offline system that ranked second in Track 2. It combines a paragraph-aware joint model, a complementary DeBERTa-large classifier, class-wise logit adjustment, and neural-BM25 evidence fusion. The ablations show that pure retrieved context is preferable to forced abstract/conclusion context, model scale improves label prediction, and the learned relation representation contributes substantially on the fixed OOF split. The oracle analyses clarify that better candidate paragraphs barely change label Macro-F1 because top-20 recall is already high, whereas selecting the correct evidence within those candidates remains a bottleneck for Joint@3. Future work should use document-grouped evaluation, repeated seeds, learned long-document retrieval, and exact match-empty post-processing.

\paragraph{Disclosure of Interests.} The authors have no competing interests to declare that are relevant to the content of this article.

\paragraph{Data and Code Availability.} The experiments use the organizer-provided NLPCC 2026 Task 10 data. The public team repository (\url{https://github.com/yanlingli031205-wq/Task10-Team-Lenormand}, default branch: \texttt{final}) contains the training and inference scripts, fixed fold identifiers, notebooks, submitted prediction archives, intermediate OOF outputs, and supplementary-analysis code and results. Organizer-provided JSONL data are excluded by the repository policy. Model weights are not committed, and exact dependency versions were not locked in the original Colab environment; the repository therefore supports pipeline inspection and re-execution when the task data and checkpoints are supplied, but is not a self-contained bitwise reproduction package.

\clearpage

\bibliographystyle{plain}
\bibliography{references}

\end{document}